\documentclass{article}

\usepackage{arxiv}

\usepackage[utf8]{inputenc} 
\usepackage[english]{babel}
\usepackage{csquotes}
\usepackage[T1]{fontenc}    
\usepackage[hidelinks]{hyperref}       
\usepackage{url}            
\usepackage{booktabs}       
\usepackage{tabularx}
\usepackage{amsmath,amssymb,amsfonts,amsthm}       
\usepackage{mathtools}
\usepackage{nicefrac}       
\usepackage{microtype}      
\usepackage{graphicx}
\usepackage{doi}
\usepackage{tikz}
\usepackage{pgfplots} 
\usepackage{pgfgantt}
\usepackage{pdflscape}
\pgfplotsset{compat=newest} 
\pgfplotsset{plot coordinates/math parser=false}

\usepackage[style=authoryear,backend=bibtex,natbib=true,url=false,doi=false
,isbn=false,date=year,maxbibnames=10,minbibnames=10,maxcitenames=2,mincitenames=1,giveninits=true
]{biblatex}
\bibliography{./sources}

\title{On Dissipativity of Cross-Entropy Loss in Training ResNets} 

\author{Jens~Püttschneider \\
	Institute of Energy Systems, Energy Efficiency\\
	and Energy Economics\\
	TU Dortmund University\\
	44227 Dortmund, Germany \\
	\texttt{jens.puettschneider@tu-dortmund.de} \\
	\And
	Timm~Faulwasser \\
	Institute of Control Systems \\
	Hamburg University of Technology \\
	21073 Hamburg, Germany \\
	\texttt{timm.faulwasser@ieee.org} \\
}

\renewcommand{\shorttitle}{On Dissipativity of Cross-Entropy Loss in Training ResNets}

\hypersetup{
	pdftitle={On Dissipativity of Cross-Entropy Loss in Training ResNets},
	pdfsubject={},
	pdfauthor={Jens~Püttschneider, Timm~Faulwasser},
	pdfkeywords={},
}

\newcommand{\End}{\hfill $\square$}
\newcommand{\PfEnd}{\hfill $\blacksquare$}

\newcounter{commoncounter}
\newtheorem{assum}[commoncounter]{Assumption}
\newtheorem{defn}[commoncounter]{Definition}
\newtheorem{lem}[commoncounter]{Lemma}
\newtheorem{cor}[commoncounter]{Corollary}
\newtheorem{prop}[commoncounter]{Proposition}
\newtheorem{rem}[commoncounter]{Remark}
\newenvironment{pf}{\begin{proof}}{\end{proof}}

\newlength{\figurebasewidth}

\newlength\figureheight
\newlength\figurewidth

\newcommand{\labelsymb}{y}
\newcommand{\statesymb}{x}
\newcommand{\inputsymb}{u}
\newcommand{\stateinputsymb}{z}

\newcommand{\stateset}{\mathbb{X}}
\newcommand{\stateinputset}{\mathbb{Z}}
\newcommand{\labelset}{\mathbb{Y}}

\newcommand{\labelensemblesymb}{\mathbf{\labelsymb}}
\newcommand{\stateensemblesymb}{\mathbf{\statesymb}}
\newcommand{\inputensemblesymb}{\inputsymb}
\newcommand{\stateinputensemblesymb}{\mathbf{\stateinputsymb}}

\newcommand{\projstatesymb}{\tilde{\statesymb}}
\newcommand{\projminimizer}{\tilde{\statesymb}^\star}

\newcommand{\inputensembleinfinitesymb}{\tilde{\inputensemblesymb}}

\newcommand{\labelensembleinitialsymb}{\labelensemblesymb}
\newcommand{\stateensembleinitialsymb}{\stateensemblesymb^0}

\newcommand{\datasetsize}{D}

\newcommand{\resnetensembledyn}{\mathbf{f}_{\text{d}}}

\newcommand{\dist}{\mathrm{dist}}
\newcommand{\lossfunc}{\ell_{\mathrm{f}}}
\newcommand{\softlossfunc}{\tilde{\ell}_{\mathrm{f}}}
\newcommand{\softlossfuncS}{\tilde{\ell}_{\mathrm{f,s}}}
\newcommand{\softlossmin}{\tilde{\ell}^{\star}}
\newcommand{\softlabel}{\tilde{q}}

\newcommand{\offsetlossfunc}{\tilde{\ell}_{\mathrm{f}}}
\newcommand{\offsetsoftlossfunc}{\tilde{\ell}_{\mathrm{f}}}

\newcommand{\numclassessymb}{C}
\newcommand{\datapointdimsymb}{C}
\newcommand{\networkdepthsymb}{N}

\newcommand{\minimizerset}{\mathbb{X}^\star}

\newcommand{\steadystateinputset}{\bar{\mathbb{Z}}}
\newcommand{\optimalstateset}{\mathbb{X}^\star}

\newcommand{\optimalsteadystateset}{\bar{\mathbb{X}}^\star}

\newcommand{\optimalsteadystateinputset}{\bar{\mathbb{Z}}^\star}

\newcommand{\optimalsteadystate}{\bar{\stateensemblesymb}^\star}
\newcommand{\optimalsteadyinput}{\bar{\inputensemblesymb}^\star}
\DeclareMathOperator*{\argmax}{arg\,max}
\DeclareMathOperator*{\argmin}{arg\,min}

\newcommand{\storagefuncresnet}{\lambda}

\newcommand{\vectone}[1]{\mathbf{1}^{#1}}
\newcommand{\identitymat}[1]{I^{#1}}

\newcommand{\vectoneC}{\vectone{C}}
\newcommand{\vectoneCtop}{\mathbf{1}^{C\top}}
\newcommand{\identitymatC}{\identitymat{C}}

\newcommand{\vectoridx}[2]{\left[{#1}\right]_{#2}}

\begin{document}
	\setlength{\figurebasewidth}{{0.50\columnwidth}}
	\maketitle
	
	\begin{abstract}
The training of ResNets and neural ODEs can be formulated and analyzed from the perspective of optimal control. This paper proposes a dissipative formulation of the training of ResNets and neural ODEs for classification problems by including a variant of the cross-entropy as a regularization in the stage cost. Based on the dissipative formulation of the training, we prove that the trained ResNet exhibit the turnpike phenomenon. We then illustrate that the training exhibits the turnpike phenomenon by training on the two spirals and MNIST datasets. This can be used to find very shallow networks suitable for a given
classification task.

\end{abstract}

	\keywords{Optimal control\and Dissipativity\and Deep learning\and Neural networks\and Turnpike property }
	
	\section{Introduction}
\label{sec:introduction}

Deep learning (DL) and (optimal) control theory share many interesting connections. 
For example, the backpropagation  in neural network training appears in optimal control as adjoint sensitivity equation \citep{ESTEVEYAGUE2023105452,faulwasser2021turnpike,esteve2021largetime}. 

Moreover, the training of Neural Networks (NNs) with constant width in each layer can be formulated as an Optimal Control Problem (OCP) \citep{li2017maximum,esteve2021largetime}.
In this context, the layer-to-layer propagation of the data is considered a dynamical system on a finite horizon corresponding to the depth of the network. 
The solution to the OCP determines the weights and biases of the neural network, i.e.,  weights and biases are the control inputs to drive the data to a desired point in the terminal layer determined by the label and loss function.
In particular, the system and control perspective is helpful for Residual Neural Networks (ResNets) \citep{he2016deep}, which can be regarded as Euler forward discretizations of neural ODEs \citep{NEURIPS2018_69386f6b}.
 \cite{chang2018reversible} analyze the reversibility and stability of the ResNet dynamics based on their continuous time counterparts.

\cite{esteve2021largetime} and \cite{faulwasser2021turnpike} have suggested to include a regularization term based on the states of the hidden layers  in the training OCP.  
\cite{faulwasser2021turnpike} analyze the dissipativity and the related turnpike property from an optimal control point of view when utilizing a quadratic ($\ell_2$) stage cost regularization.  They provide constructive depth bounds in this setting.
\cite{esteve2021largetime} investigate turnpikes in the training of neural ODEs when using a stage cost based on the OCP optimality conditions.
Morevover, \cite{RuizBalet2021} establish reachability properties for neural ODEs with the ReLU activation function.

The role of dissipativity in deep learning is also analyzed by \cite{feng2011stability,zeng2015stability} for cellular neural networks, by \cite{revay2023recurrent} for recurrent equilibrium networks, and by \cite{martinelli2023unconstrained} for neural ODEs.
Moreover, passivity properties, like the ones in Hamiltonian NNs, are used to address vanishing and exploding gradients during training \citep{galimberti2023hamiltonian}.

 In a nutshell, the turnpike phenomenon is related to similarity properties of optimal solutions for varying initial condition and varying horizon length. The concept originated in optimal control approaches to economics~\citep{Dorfman58,Mckenzie76} and early observations are due to \cite{Ramsey28} and \cite{vonNeumann38}. 
In the analysis of turnpike properties in optimal control there has been recent progress along two avenues: analysis of the optimality system \citep{Trelat15a,sakamoto2021turnpike} and leveraging dissipativity properties of the OCP~\citep{grune2016relation,epfl:faulwasser15h,damm2014exponential}. Interestingly the dissipativity route is linked to the foundational work of \cite{Willems71a} on infinite-horizon least-squares optimal control but it also generalizes to non-quadratic objectives and nonlinear systems~\citep{tudo:faulwasser21a}. The turnpike can be regarded as the attractor of the infinite horizon optimal solutions~\citep{trelat2023linear,tudo:faulwasser21a}. We refer to \cite{grune2022dissipativity} and to \cite{tudo:faulwasser22a} for recent literature overviews. A recent trend in dissipativity-based analysis of OCPs is the generalization of turnpike properties towards more general turnpike objects such as subspaces~\citep{tudo:schaller21a} or manifolds~\citep{tudo:faulwasser22e,karsai2024manifold}. 

In this paper, we leverage subspace turnpike concepts to analyze ResNet training from an optimal control perspective. 
In contrast to the quadratic regularization costs proposed by \cite{esteve2021largetime} and \cite{faulwasser2021turnpike}, we consider a variant of the cross-entropy for classification tasks to obtain a dissipative formulation of ResNet training for classification tasks.
Specifically, using the cross-entropy with soft labels we derive locally a suitable quadratic lower bound to the loss function. We also show that the soft cross-entropy behaves similar to \emph{Huber loss} \citep{huber1992robust}, i.e. locally quadratic with linear asymptotics. This lower bound can be used to prove the strict dissipativity of our training OCP formulation with respect to the linear subspace of steady minimizers of the soft cross-entropy. We prove the existence of subspace turnpikes in the underlying training problem. Moreover, we extend our result to continuous-time formulations with neural ODEs and we propose sufficient conditions which enable extension to other NN architectures derives via implicit or explicit discretization of neural ODEs.

The remainder of the paper is structured as follows: Section \ref{sec:background} introduces the optimal control formulation of deep learning. Section \ref{sec:resnets} provides the dissipative formulation of neural network training. Section \ref{sec:tp} then uses the dissipative formulation to prove the existence of turnpikes in the trained NN. 
Section \ref{sec:neuralODEs} extends the dissipative training formulation to neural ODEs.
Section \ref{sec:experiments} validates the formulation by training a ResNet on the two spirals dataset and MNIST. We end with a conclusion and an outlook in Section \ref{sec:conclusion}.
	\section{Optimal Control and ResNet Training}
\label{sec:background}
 
The training of Neural Networks (NN) and neural ODEs in deep learning can be cast as an optimal control problem, where the control inputs are the network parameters that steer the data through the layers towards a representation in the last layer corresponding to the label. 
The perspective of optimal control is particularly beneficial for Residual Networks (ResNets), which can be interpreted as Euler forward discretizations of neural ODEs.

The propagation of a data point $\statesymb^i$ through the layers of a ResNet can be conceptualized as a discrete time dynamical system of the form
\begin{equation}
	\statesymb^i_{k+1} = \statesymb^i_k + \sigma\left(A_k \statesymb^i_k+b_k\right), \quad \statesymb^i_0=\statesymb^i \in \mathbb{R}^{\datapointdimsymb},
	\label{eq:resnet}
\end{equation}
where the time step $k \in \mathbb{N}_{[0,\networkdepthsymb-1]}$ corresponds to the index of the residual layer in the $\networkdepthsymb$-layer network.
When training the ResNet, the parameters, weights $A_k$ and biases $b_k$, are optimized to best fit the training data. 
Throughout this paper, the scalar and continuous activation function $\sigma: \mathbb{R} \rightarrow \mathbb{R}$ is applied element-wise to the activation vector $A_k \statesymb^i_k+b_k$. Moreover, we require $\sigma(0)=0$ such that ResNets can render each state $x$ a steady state by choosing $A=0$ and $b=0$. This implies ResNets can learn identity mappings, preserving information from one layer to the next and thus often show superior performance of deeper ResNets over shallower ones \citep{he2016deep}.
\footnote{In this paper, we refer to the full dynamics \eqref{eq:resnet}  evaluated at some step $k$ as \textit{residual layer}. However, in the machine layer literature the term \textit{residual block} is also used, whereby the notion \textit{layer} is reserved for the mapping $x_k \mapsto A_k \statesymb^i_k+b_k$ \citep{he2016deep}.}

The data propagated through the network corresponds to the state $\statesymb^i_k$, its initial condition is a feature $\statesymb^i$ from the dataset
\begin{equation*}
	\mathbb{D}=\left\{\left(\statesymb^1,\labelsymb^1\right), \dots, \left(\statesymb^{\datasetsize},\labelsymb^{\datasetsize}\right)\right\}.
\end{equation*}
The label $y^i$ determines the target of the state propagation.
For classification task with $\numclassessymb$ different classes, the label $\labelsymb^i  \in \labelset =\mathbb{N}_{[1, \numclassessymb]}$ represents the class index.\footnote{\label{foot:outputlayer}One could formally introduce an output equation in \eqref{eq:resnet} which would resemble the functionality of an output layer in the ResNet. The parameters of the output layer can also be optimized for in the training process. Which, in the language of optimal control, leads to additional degrees of freedom.  In view of our later analysis in Section \ref{sec:resnets} we simplify the exposition by not explicitly detailing the output layer. One could, however, include this by considering explicit dependence of the loss function on decision parameters.}
The selection of hyperparameters that define the neural network architecture, such as its depths $\networkdepthsymb \in \mathbb{N}$ and activation function $\sigma(\cdot)$ is a crucial design decision that aims at finding the neural network structure best suited for the given training problem.

Closely related to ResNets are neural ODEs that provide a continuous-time formulation of deep learning
\begin{equation} \label{eq:NODE}
	\dot{\statesymb}^i(t) = \sigma \left(A(t)x^i(t)+b(t)\right), \quad \statesymb^i(0) = \statesymb^i \in \mathbb{R}^{\datapointdimsymb},
\end{equation}
for a suitable discretization of the continuous parameters $A(t)$ and $b(t)$. In particular, fixed-step size Euler forward discretization of the neural ODE above yields the ResNet from \eqref{eq:resnet}.

\subsection{Optimal Control Formulation of Deep Learning}

To formalize the training on the entire dataset, we stack the individual data points 
\begin{equation}
	\stateensembleinitialsymb\dot{=}\left(\statesymb^{1\top}, \hdots, \statesymb^{\datasetsize\top}\right)^\top 
	\quad 
	\labelensembleinitialsymb \dot{=}\left(\labelsymb^1, \hdots, \labelsymb^{\datasetsize}\right)^\top,
	\label{eq:stacked_dataset}
\end{equation}
which allows writing the data as $\mathbb{D}=\{(\stateensembleinitialsymb, \labelensembleinitialsymb )\}$.
For the stacked data, the ResNet dynamics are
\begin{equation}
	\begin{aligned}
		\stateensemblesymb_{k+1} &= \stateensemblesymb_k + \sigma\left((\identitymat{\datasetsize} \otimes A_k)\stateensemblesymb_k+  (\vectone{\datasetsize} \otimes b_k) \right)  \\
		&\coloneqq \resnetensembledyn(\stateensemblesymb_k, \inputensemblesymb_k), \qquad \qquad \stateensemblesymb_0=\stateensembleinitialsymb \in \mathbb{R}^{\datasetsize \cdot \datapointdimsymb},
	\end{aligned}	
	\label{eq:ensemble_dynamics}
\end{equation}
where $\inputensemblesymb_k=(\text{vect}(A_k)^\top,b_k^\top)^\top \in \mathbb{R}^{n_u}$ is the vectorized input of layer $k$ with dimensionality $n_u=\datapointdimsymb^2 + \datapointdimsymb$, $I^\datasetsize \in\mathbb{R}^{\datasetsize \times \datasetsize}$ is the identity matrix, $1^\datasetsize \in \mathbb{R}^\datasetsize$ the vector of all ones, and $\otimes$ refers to the Kronecker product.
Note that the boldface state variable $\stateensemblesymb$ indicates the stacked state for all data samples, which are in this formulation  simultaneously controlled by one   ResNet, i.e., one input sequence $\inputensemblesymb_k$ which entails the weights and biases.
The scalar $\gamma$ is used to trade-off the importance of the regularization against the loss function at the terminal layer.
As a shorthand, we write state-input pairs as $\stateinputensemblesymb=(\stateensemblesymb, \inputensemblesymb)$.

For a number of residual layers $\networkdepthsymb$ (i.e. for a network depth N), training the ResNet can be formulated as the discrete-time OCP
\begin{subequations} \label{eq:training_ocp}
\begin{align}
		V^{\gamma}_N\left(\stateensembleinitialsymb\right) = 
		\min_{\inputensemblesymb_{0,\dots, \networkdepthsymb-1}}& \sum_{k=0}^{\networkdepthsymb-1} \ell(\stateensemblesymb_k, \inputensemblesymb_k) + \gamma\lossfunc(\stateensemblesymb_N, \labelensemblesymb)\\
		\text{s.t.} & ~ \forall k \in \mathbb{N}_{[0,N-1]}\notag \\
		\quad \quad& \stateensemblesymb_{k+1} = \resnetensembledyn(\stateensemblesymb_k, \inputensemblesymb_k),   \label{eq:training_ocp_dyn}\\
		& \stateensemblesymb_0 = \stateensembleinitialsymb \in \mathbb{R}^{\datapointdimsymb \cdot \datasetsize},
	\end{align}
\end{subequations}
where the Mayer term  (terminal penalty) $\lossfunc$ is the loss function applied to the entire data set and describes the quality of the NN output. 
Typically one uses the empirical loss, averaging the loss over the dataset for which, 
with slight abuse of notation,
\begin{equation*}
	\lossfunc(\stateensemblesymb_\networkdepthsymb, \labelensemblesymb) \doteq \frac{1}{\datasetsize}\sum_{i=1}^{\datasetsize} \lossfunc(\statesymb_N(x^i), \labelsymb^i),
\end{equation*}
 the boldface letter arguments $\stateensemblesymb$ and $\labelensemblesymb$ indicate the dependence on the entire dataset, whereas $\lossfunc(\statesymb_N(x^i), \labelsymb^i)$ is the loss of the individual data sample $(\statesymb^i, \labelsymb^i)$.
Additionally, the stage cost $\ell:\mathbb{R}^{\datapointdimsymb \cdot \datasetsize} \times  \mathbb{R}^{n_u} \rightarrow \mathbb{R}^+_0$ captures the regularization terms;  particular choices of the stage cost will be introduced later.

The solution to OCP \eqref{eq:training_ocp} are the NN parameters, weights and biases, i.e. the control inputs denoted as $\inputensemblesymb^\star(\stateensembleinitialsymb)$ and the resulting ensemble data trajectories $\mathbf{x}^\star\left(\stateensembleinitialsymb\right)$, which depend on the data set as highlighted by $\stateensembleinitialsymb$.
From the optimal control perspective, the main difficulty of NN training lies in the simultaneous control of $\datasetsize$ data samples with only one network, i.e. only one control signal is applied to many initial conditions.

\subsection{Cross-Entropy Loss for Classification}
In classification, the goal is to predict a discrete class $y$ given the feature $x$ \citep{lecun2015deep}. For the loss function this means comparing the continuous state to the discrete class label. This is typically done by the cross-entropy loss function first proposed by \cite{cox1958regression}. The cross-entropy first calculates probabilities for all possible classes $y \in \labelset$ from the NN output state $x$ using the softmax activation function 
\begin{equation*}
	p(\labelsymb|x) = \frac{e^{\vectoridx{\statesymb}{\labelsymb}}}{\sum_{i=1}^{\numclassessymb} e^{\vectoridx{\statesymb}{i}}}, 
\end{equation*}
whereby the operator $\vectoridx{\cdot}{i}$ accesses the $i$-th component of a vector.
The predicted class $\hat{\labelsymb} = \argmax_i 	p(i|\statesymb)$ is the class with the highest probability.
The probabilities for all classes are then arranged into the probability vector
\begin{equation}
	p(x) = \begin{bmatrix}
		p(1|x) , ..., p(C|x)
	\end{bmatrix}^\top.
	\label{eq:softmax_probability_vector}
\end{equation} 
Likewise to \eqref{eq:softmax_probability_vector}, we define the vector of target probabilities $q(y)$ determined by the label $y$.
Typically binary targets $q(i|y) = \delta_{y,i}$ are used, where the probability of the labeled class $y$ is one and zero for all other classes.

Then, the output probability distribution vector $p(x)$ is compared to the target vector $q(y)$ induced by the labels using the cross-entropy
\begin{equation*}
	\lossfunc(\statesymb,\labelsymb) =- H(p(x), q(y))) = -\sum_{i = 1}^{C} q(i|y)\log p(i|x).
\end{equation*}
For the binary targets, the cross-entropy only depends on the softmax probability of the correct class
\begin{equation}
	\lossfunc(\statesymb,\labelsymb) =-\log p(y|x).
	\label{eq:cross_entropy}
\end{equation}

\subsection{Dissipativity of OCPs} 
We recall the definition of dissipativity of OCPs introduced by \cite{Angeli2012}, based on the dissipativity notion for open dynamical systems 
coined by \cite{Willems1972}. Moreover, we use an extended definition of dissipatvity with respect to a set of optimal steady state pairs $\optimalsteadystateinputset$, similar to \cite{8814305,Mueller2021}.
The set of steady state pairs is given by 
\begin{equation} \label{eq:Xs}
\bar{\stateinputset} = \left\{\bar{\stateinputensemblesymb}=(\bar{\stateensemblesymb}, \bar{\inputensemblesymb}) \in\mathbb{R}^{\datasetsize\cdot \numclassessymb}\times \mathbb{R}^{(C^2+C)N} \,|\,\bar{\stateensemblesymb} = \resnetensembledyn(\bar{\stateensemblesymb},\bar{\inputensemblesymb})
\right\}
\end{equation}
	Optimal steady state pairs are computed via
	\begin{equation}
		\begin{aligned}
			\bar{\mathbf{z}}^\star \in
			\argmin_{\bar{\mathbf{z}} } 
			&\; \ell(\bar{\mathbf{z}}) \quad 
			 \text{s.t.} \quad \bar{\mathbf z} \in \steadystateinputset 
		\end{aligned}
	\end{equation}
	and the set of all optimal steady states is written as $\optimalsteadystateinputset \subseteq  \steadystateinputset$.

Recall that the distance between a point $x\in \mathbb{R}^{n_x}$ and the closed set $\mathbb{X} \subset \mathbb{R}^{n_x}$ is given by
		\[
			\mathrm{dist}\left(x,  \mathbb{X}\right) \doteq \min_{x^\prime\in \mathbb{X}} \lVert x - x^\prime\rVert.
		\]

 \begin{defn}[Strict dissipativity  in discrete time] \label{def:DI}
	\begin{subequations} \label{eq:dissipation_inequality}
		The discrete time dynamical system \eqref{eq:ensemble_dynamics} is said to be dissipative with respect to a set of steady-state pairs $\optimalsteadystateinputset$ if there exists a non-negative storage function $\lambda: \mathbb{R}^{\datasetsize\cdot \numclassessymb} \rightarrow \mathbb{R}^+_0$ such that for all $\stateinputensemblesymb=(\stateensemblesymb,\inputensemblesymb)$ and all $\bar{\stateinputensemblesymb}^\star \in \optimalsteadystateinputset$ 
		\begin{equation}
			\lambda(\resnetensembledyn(\stateinputensemblesymb)) - \lambda(\stateensemblesymb) \leq \ell(\stateinputensemblesymb) - \ell(\bar{\stateinputensemblesymb}^\star).	
		\end{equation}
		\label{enum:strict_dissipativity:dissipativity_dynamical_sys}
		If additionally, there exists $\alpha_{\ell}  \in \mathcal{K}$ such that 
		\begin{equation}
			\lambda(\resnetensembledyn(\stateinputensemblesymb)) - \lambda(\stateensemblesymb) \leq \ell(\stateinputensemblesymb) - \ell(\bar{\stateinputensemblesymb}^\star).
			- \alpha_{\ell}\left(
			\mathrm{dist}\left(\stateinputensemblesymb, \optimalsteadystateinputset\right)\right),
			\label{eq:strict_dissipation_inequality}
		\end{equation}
		then the system \eqref{eq:ensemble_dynamics} is said to be strictly $\stateensemblesymb-\inputensemblesymb$ dissipative with respect to $\optimalsteadystateinputset$ and for $\stateinputensemblesymb$ replaced by $\stateensemblesymb$ in the class $\mathcal K$ function the system is said to be strictly $\stateensemblesymb$ dissipative. 
	
		The OCP \eqref{eq:training_ocp} is said to be (strictly) $\stateensemblesymb-\inputensemblesymb$ dissipative with respect to $\optimalsteadystateinputset$ if for all $N\in\mathbb{N}$ and all $\stateensemblesymb_0 \in \mathbf{X}_0$, the dissipation inequalities \eqref{eq:dissipation_inequality} hold along any optimal trajectory of \eqref{eq:training_ocp}.
	\end{subequations}
	\End
\end{defn}

For a singleton set $\optimalsteadystateinputset = \left\{ \optimalsteadystate; \optimalsteadyinput \right\}$ the above definition corresponds to the standard  dissipativity notion with respect to an optimal steady state $\optimalsteadystate$.
Notice that, in view of Definition \ref{def:DI}, the dissipativity of OCP \eqref{eq:training_ocp} only depends on the regularization $\ell(\stateensemblesymb,\inputensemblesymb)$ and not on the loss function, i.e. the Mayer term $\ell_{\mathrm f}$.

	\section{Dissipativity  of Cross-Entropy Loss in ResNets}
\label{sec:resnets}
Next, we turn towards analyzing the dissipativity properties of OCP~\eqref{eq:training_ocp}. As a preparatory steps we investigate the minimization properties of cross-entropy and its variant the soft cross-entropy.

\subsection{Conceptual Difficulties of Standard Cross-Entropy}
Consider the stage cost 
	\begin{equation}
		\ell(\stateensemblesymb, \inputensemblesymb) = \lossfunc(\stateensemblesymb, \labelensemblesymb) + r \lVert \inputensemblesymb \rVert^2,
		\label{eq:stagecost}
	\end{equation} 
	in the training OCP \eqref{eq:training_ocp}. 
\begin{lem}[No minimizers  for cross-entropy]~\\
The stage cost \eqref{eq:stagecost} has no minimizers in $\mathbb{R}^{C\cdot D}$.
\end{lem}
\begin{pf}
Minimizing the loss function $\lossfunc$ implies that for each sample $\left(x^i, y^i\right)$ from the dataset the softmax probability of the labelled class is one
		\[p(\labelsymb^i|x^i) = \frac{e^{\vectoridx{\statesymb^i}{\labelsymb^i}}}{\sum_{j=1}^{\numclassessymb} e^{\vectoridx{\statesymb^i}{j}}}=1.\] 
		Hence we have $\vectoridx{x^i}{j} \rightarrow \infty$ if $j = \labelsymb_i$ and $\vectoridx{x^i}{j} \rightarrow -\infty$ if $j \neq \labelsymb_i$. In other words, the optimal steady states are pushed to infinity, while the infimum of the loss function is $\ell(z^\star) =0$.\PfEnd
\end{pf}
This result implies that the strict dissipation inequality \eqref{eq:strict_dissipation_inequality} cannot hold with $\alpha_\ell \in \mathcal K_\infty$, i.e., \eqref{eq:strict_dissipation_inequality} does not hold if $\alpha_\ell$ is radially unbounded. Another consequence of minimizers pushed to infinity is that classic turnpike analysis concepts, which rely on reachability assumptions break down and that the time the optimal solutions can spend far away from the optimal steady state is not bounded independent of the horizon length. Indeed, for any finite horizon in OCP \eqref{eq:training_ocp} the solutions are always   infinitely far away from the optimal steady state. As we see later this contradicts the turnpike property as used in Section~\ref{sec:tp}.	
Subsequently, instead of changing the turnpike concepts and as NN of finite depth are more application relevant, we adapt the considered loss function.

\subsection{Soft Cross-Entropy and Its Properties}
Label smoothing, first introduced by \cite{7780677}, utilizes a target probability reachable by the softmax activation function. For a sample from class $y$, the target probability for class $i$ of 
\begin{equation}
	\softlabel(i|\labelsymb) = 
	\begin{cases}
		p_d \quad & i = \labelsymb \\
		\frac{1-p_d}{C-1}  \quad & i \neq \labelsymb
	\end{cases},
	\label{eq:soft_target}
\end{equation}
is used.
The main motivation of the soft cross-entropy is robustification of classification tasks with label noise, i.e. uncertainties in the labels. These occur if some of the labels observed from the dataset do not match the actual label, this can be due to human errors in the labeling process or due to ambiguity in the classification task itself, e.g., multiple classes are present in the same image. 
In these situations the soft cross-entropy has produced better generalization properties in empirical studies \citep{largemargindnns}. 
To prevent overconfident false classification on these examples, the true class is assigned a probability $p_d$ of close to one, while remaining non-zero probability is shared uniformly between the $C-1$ incorrect classes.

The soft cross-entropy, the cross-entropy between the softmax probabilities $p(x)$ and the smoothed target distribution $\softlabel(y)$, is then used as the loss
\begin{equation}
	\begin{split}
		\softlossfunc(x,y) &= -H(p(x),\softlabel(y)) -\softlossmin \\
		&=  - \sum_{\substack{i=1}}^{C}  \softlabel(i|y)\log p(i|x) -\softlossmin.
	\end{split}
	\label{eq:soft_cross_entropy}
\end{equation}
In this work, we use the constant offset $\softlossmin=- H(\softlabel(y),\softlabel(y))$, such that the minimal value becomes zero.
To analyze the dissipativity of the training using the soft cross-entropy we need to find its stationary minimizer.

\begin{lem}[Minimizers  of  soft cross-entropy]	\label{lemma:minimization_cross_entropy}
	Consider the minimization of the soft cross-entropy \eqref{eq:soft_cross_entropy} for the label $y$ 
	\begin{equation*}
		\minimizerset_{\labelsymb}=\argmin_{\statesymb_\labelsymb \in \mathbb{R}^{C}} \quad \softlossfunc(\statesymb_\labelsymb,\labelsymb).
	\end{equation*}
	Its minimizers form the line
	\begin{equation*}
		\minimizerset_{\labelsymb} = 	\left\{x\in \mathbb{R}^C  \middle|\vectoridx{x}{c} = \vectoridx{x}{y} + \delta, \: \forall c \in \labelset \setminus \left\{ \labelsymb \right\}, \: \forall \vectoridx{x}{y} \in \mathbb{R},   \right\},
		\label{eq:minimizing_network_states}
	\end{equation*}
	where $\delta =  -\log\left(\frac{(C-1)p_d}{1-p_d}\right)$.
	The corresponding minimum value
	\begin{equation*}
		\min_{x} \quad \softlossfunc(\statesymb,\labelsymb) = \softlossmin - \softlossmin = 0,
	\end{equation*}
	is independent of the label $y$.
	\End
	\end{lem}
	\begin{pf}
		We restrict the proof to the case of class $\labelsymb=1$ without loss of generality, since the categorical cross-entropy is symmetric to exchanging label $\labelsymb$ and $\labelsymb'$ with the exchange of state $\statesymb_\labelsymb$ and $\statesymb_{\labelsymb'}$.
		The soft cross-entropy is the cross-entropy of the softmax probabilities $p(x)$ and the smoothed target probabilities $\softlabel(y)$
		$\softlossfunc(x,y) = -H(p(x),\softlabel(y))-\softlossmin$. It is minimal if the vectors of softmax and target probabilities match,i.e., if $p(x)=\softlabel(y)$.		
		For class $\labelsymb=1$, $p(\statesymb)=\softlabel(y)$ implies
		\newcommand{\classymb}{c}
		\begin{equation*}
			p(\classymb|x^\star) = \frac{e^{\vectoridx{x^\star}{\classymb}}}{\sum_{i=1}^{C} e^{\vectoridx{x^\star}{i}}} =	
			\begin{cases}
				p_d &\quad \classymb = y \\
				\frac{1-p_d}{C-1} &\quad \classymb \neq y.
			\end{cases}
		\end{equation*}
		The equations for the incorrect classes, $p(\classymb \neq \labelsymb| \statesymb)$, only differ in the numerator $e^{\vectoridx{x^\star}{\classymb \neq \labelsymb}}$, such that all $\vectoridx{x^\star}{\classymb \neq \labelsymb}$ have to be equal.
		Dividing $p(\labelsymb|\statesymb)$ by $p(c\neq\labelsymb|\statesymb)$ gives $\frac{e^{\vectoridx{x^\star}{\labelsymb}}}{e^{\vectoridx{x^\star}{c\neq \labelsymb}}} = \frac{(\numclassessymb-1)p_d}{1-p_d}$.
		We obtain the line, $\forall c \in \labelset \setminus \left\{ y \right\}, \forall \vectoridx{x^\star}{\labelsymb} \in \mathbb{R}$ 
		\[\vectoridx{x^\star}{\classymb}  =  \vectoridx{x^\star}{\labelsymb} -\log\left(\frac{(\numclassessymb-1)p_d}{1-p_d}\right)\] which forms $\minimizerset_{\labelsymb}$.
		\PfEnd
	\end{pf}

\begin{rem}[Large data  with $\dim x > C$]
	\label{remark:n_new_c}
	The softmax probabilities and therefore the loss function, consider a state vector with one state component per class, $\dim x=C$. 
	If a state with $\dim x=n > C$ is used, only the $C$-first (or $C$ specified) components of $x$ are considered in the loss function.
	Then, the remaining components of the state $\vectoridx{x}{C+1, ..., n}$ do not contribute to the loss nor the distance to $\minimizerset_{\labelsymb}$.
	Without loss of generality, our analysis is done only for the case that the dimension of the state matches the number of classes $\dim x = C$. Indeed, our main dissipation and turnpike results leverage the geometry of the softmax cross entropy and thus also hold for the case that $n> C$.
	
	Moreover, an output layer could be included to map $x$ with $\dim x = n$ to an output $y_\mathrm{p}$ predicting the label $y$ with $\dim y_\mathrm{p} = C$. This may potentially induce additional decision variable in the training problem, cf. Footnote \ref{foot:outputlayer}.	
	\End
\end{rem}
Now, we introduce several technical lemmas relating $\offsetlossfunc(\statesymb,\labelsymb)$ to the distance to the set of it minimizers $\minimizerset_{\labelsymb}$. This paves the road to prove dissipativity.

\begin{lem}[Invariance of soft cross-entropy]
	\label{lemma:translational_invariance_softmax_ce}
	The softmax probabilities and the soft cross-entropy are invariant to the transformation $T: \stateset \rightarrow \stateset$ of the form
	\begin{equation}
		T = \identitymatC - \frac{1}{C} \vectoneC \vectoneCtop,
		\label{eq:invarianttransformations}
	\end{equation}
	where $\identitymatC \in \mathbb{R}^{\numclassessymb \times \numclassessymb}$ is the identity matrix, $\vectoneC \in \mathbb{R}^\numclassessymb$ denotes the the vector of all ones.
	This transformation subtracts the average of all components from all components.
	\End
	\end{lem}
	\begin{pf}
		For all classes $\labelsymb = 1,\dots, \numclassessymb$, the translated softmax probability is invariant to \eqref{eq:invarianttransformations}, as
		\begin{equation*}
			p(\labelsymb|T \statesymb) = \frac{e^{x_\labelsymb - x_{\mathrm{avg}}}}{\sum_{j=1}^{C} e^{x_j- x_{\mathrm{avg}}}} = \frac{e^{x_\labelsymb}}{\sum_{j=1}^{C} e^{x_j}} = p(\labelsymb|x),
		\end{equation*}
		where $x_{\mathrm{avg}} = \dfrac{1}{C} \vectoneCtop x$ is the average of all components of $x$.
		The soft cross-entropy is a function of the invariant sofmax probabilities and therefore also invariant to the transformations \eqref{eq:invarianttransformations}.
	\PfEnd
	\end{pf}

Due	to the translational invariance of the classification of Lemma \ref{lemma:translational_invariance_softmax_ce}, we further restrict the dissipativity analysis on the subspace $\stateset_{T}$ where $\vectoneCtop x=0$:
\begin{equation*}
	\stateset_{T} \doteq \left\{ x\in \mathbb{R}^C | \vectoneCtop x =0\right\}.
\end{equation*}

The intersection of $\stateset_{T}$ and $\stateset_\labelsymb^\star$ is given by the point $x_\labelsymb^\star = \argmin_{\projstatesymb \in \stateset_{T}} \softlossfunc(\projstatesymb, \labelsymb)$, which is component-wise defined as
\begin{equation*}
	[\projstatesymb_\labelsymb^\star]_i = 
	\begin{cases}
		-\frac{\numclassessymb}{\numclassessymb - 1} \delta  & i = \labelsymb\\
		\frac{1}{\numclassessymb} \delta  & i \neq \labelsymb.
	\end{cases} 
\end{equation*}
The projection $T:\stateset\rightarrow\stateset_{T}$, maps a point $\statesymb \in \stateset$ to $\projstatesymb \in \stateset_{T}$, while keeping the value of the soft cross-entropy constant.
\begin{lem}[$T$ preserves the distance to $\minimizerset_{\labelsymb}$]
	\label{lemma:projection_preserves_distance}
	For all $x\in \mathbb{R}^C$ 
	\begin{equation*}
		\dist(x, \minimizerset_{\labelsymb}) = \dist(T x, \minimizerset_{\labelsymb}).
	\end{equation*}
	\End \vspace*{-7mm}
	\end{lem}
	\begin{pf}
		Without loss of generality and similar to Lemma \ref{lemma:minimization_cross_entropy}, we prove this for the case $\labelsymb=1$ .
		Let $\statesymb^\star_1$ be the closest point to $\statesymb \in \mathbb{R}^\numclassessymb$ in $\minimizerset_1$, 
		then also $\statesymb^\star_1 + \alpha \vectoneC  \in \minimizerset_1, \; \forall \alpha \in \mathbb{R}$, so $d=x- \statesymb^\star_1$,   $d\perp \vectoneC$. Therefore, $d$ lies in the eigenspace of $T$ corresponding to the eigenvalue $1$ and  $\projstatesymb - \projminimizer_1 = T(\statesymb- \statesymb^\star) = \statesymb- \statesymb^\star$. 
		\PfEnd
	\end{pf}

\begin{lem}[Convexity of the soft cross-entropy]
	\label{lemma:properties_soft_cross_entropy}
	The soft cross-entropy $\softlossfunc: \mathbb{R}^\numclassessymb \times \labelset \rightarrow \mathbb{R}^+$,  \eqref{eq:soft_cross_entropy}, is convex and strictly convex when restricted to $\softlossfunc: \stateset_{T} \times \labelset \rightarrow \mathbb{R}^+$.
	 The Hessian of \eqref{eq:soft_cross_entropy} is a positive semi-definite matrix, with $0$ as a single eigenvalue and $\vectoneC$ as the corresponding eigenvector, and all other eigenvalues are strictly positive for all $x \in \mathbb{R}^\numclassessymb$, i.e., \eqref{eq:soft_cross_entropy} satisfies the second-order sufficient conditions for convexity. 
	\End
\end{lem}
	\begin{pf}
		The Hessian matrix of the cross-entropy 
		\[
			\begin{aligned}
				H(x) = \nabla_x^2 \ell(x,y) &= \text{diag}\left(p(x)) - p(x) p(x)^\top\right)\\
				& = \text{diag}\left((p(x)\right)\left(\identitymatC - \vectoneC p(x)^\top\right)
			\end{aligned}
		\]
		is independent of the target probabilities $q(y)$ and therefore equal for the regular and soft cross-entropy $\tilde{H}(x) = \nabla_x^2 \softlossfunc(x,y)=H(x)$.
		The Hessians $H(x)$ and $\tilde{H}(x)$ are positive semi-definite matrices \citep[Theorem 2]{pmlr-v97-singla19a}.
		
		Moreover, the eigenvalue $0$ has the algebraic multiplicity one, since $\text{diag}(p(x))$ is of full rank and therefore $\text{rank}(\tilde{H}(x))=\text{rank}(\identitymatC - \vectoneC p(x)^\top)$ with
		\begin{equation*}
			\text{rank}\left(\identitymatC-\vectoneC p(x)^\top\right) \geq  \text{rank}(\identitymatC) - \text{rank}\left(\vectoneC p(x)^\top\right) = C - 1. 
		\end{equation*}
		The eigenvector of the Hessian corresponding to the eigenvalue $0$ is $\vectoneC$, since $(\identitymatC - \vectoneC p(x)^\top)\vectoneC = \vectoneC - \vectoneC = 0$ and $p(x)^\top \vectoneC = 1$. 
 		
 		For all $\statesymb \in \stateset$ and all directions $d \in \mathbb{R}^C$, the soft cross-entropy satisfies the second-order sufficient conditions for convexity
		\[
		d^\top \tilde{H}(\projstatesymb) d \geq 0,
		\]
		since all eigenvalues of the Hessian $\tilde{H}(\projstatesymb)$ are larger or equal to zero.
		Moreover, for all $\projstatesymb \in \stateset_{T}$ and in all directions $d \in \mathbb{R}^C$ with $\projstatesymb + d \in \stateset_{T}$ and $d\neq 0$, the soft cross-entropy satisfies 
		\[
		d^\top \tilde{H}(\projstatesymb) d > 0,
		\]
		since the eigenvalues of the Hessian in all directions $d \perp \vectoneC$ are strictly larger than zero. That is,  the second-order sufficient conditions for strict convexity hold. Moreover, $d \perp \vectoneC$ because $\projstatesymb \in \stateset_{T}$ and $\projstatesymb + d \in \stateset_{T}$.
	\PfEnd
	\end{pf}

\begin{lem}[Soft cross-entropy lower bound]
\label{lemma:soft_cross_entropy_distance_measure}
	For the soft cross-entropy \eqref{eq:soft_cross_entropy}, there exists $\alpha\in \mathcal{K}_\infty$ such that
	\begin{equation}
		\label{eq:soflossfunc_lower_K}
		\begin{aligned}
			\softlossfunc(x, y) \geq \alpha\left(\text{dist}(x, \minimizerset_{\labelsymb})\right).
		\end{aligned}
	\end{equation}
	\End
\end{lem}
\begin{pf}
	Due to Lemmas \ref{lemma:translational_invariance_softmax_ce} and \ref{lemma:projection_preserves_distance}, both sides of the inequality \eqref{eq:soflossfunc_lower_K} are invariant to the transformation $T_{y}$ so that we assume $x=\projstatesymb\in \mathbb{X}_T$ without loss of generality.
	Moreover, similar to Lemma \ref{lemma:minimization_cross_entropy}, we consider the case $\labelsymb = 1$ without loss of generality.

	Let $\statesymb^\star_1$ be the closest point to $\projstatesymb \in \stateset_T$ in $\minimizerset_1$, 
then also $\statesymb^\star_1 + \alpha \vectoneC  \in \minimizerset_1, \; \forall \alpha \in \mathbb{R}$, so $d=x- \statesymb^\star_1$,   $d\perp \vectoneC$ and thefore $ \statesymb^\star_1 = \projstatesymb^\star_1$.
	Then \eqref{eq:soflossfunc_lower_K} implies
	\begin{equation}
		\softlossfunc(\projstatesymb, y) \geq \alpha\left(\lVert  x - \projstatesymb^\star_1\rVert\right).
	\end{equation}
	
	Consider the half-line $x_{\mathrm{s}}: [0,\infty)\times \labelset \rightarrow \mathbb{R}^C$, $x_{\mathrm{s}}(s,\labelsymb)=\projstatesymb^\star_{\labelsymb} + d \cdot s$
	starting at the minimizer $ \projstatesymb_\labelsymb^\star$, for any direction direction $d \in \mathbb{R}^C$ with $\lVert d\rVert=1$ and $d \perp \vectoneC$, such that $\lVert x_{\mathrm{s}}(s,\labelsymb) - \projstatesymb_\labelsymb^\star \rVert = s$.
	For $s\leq1$ all directions $d$ lead to the compact set 
	\[
		\Omega = \left\{ \projstatesymb \in \stateset_T \,\middle|\, \lVert \projstatesymb - \projstatesymb^\star_1 \rVert \leq 1 \right\}.
	\]
	
	Let $\lambda_{\Omega, \text{min}} = \min_{x\in \Omega} \lambda_{min}(\tilde{H}(\projstatesymb))$ be the smallest eigenvalue of $\tilde{H}(x)$ on $\Omega$ in any direction $d \perp \vectoneC$. The function $\lambda_{\text{min}}: \mathbb{R}^{C\times C} \rightarrow \mathbb{R}$ calculates the minimal eigenvalues of the matrix $A$ and is continuous in the entries of $A$ \citep[Theorem 2.4.9]{0521386322}. Moreover, the matrix $\tilde{H}(x)$ is continuous in $x$ and therefore the function $\lambda_{\text{min}}(\tilde{H}(x))$ is continuous in $x$, such that the minimum on $\Omega$ exists and is strictly positive, i.e., $\lambda_{\Omega, \text{min}}\geq 0$.
Consequently, we have that
	\[
		\offsetlossfunc(\projstatesymb, \labelsymb) \geq \dfrac{\lambda_{\Omega, \text{min}}}{2}  \left(\tilde{x} - \projminimizer_{1}\right)^\top  \left(\tilde{x} - \projminimizer_{1}\right) = \dfrac{\lambda_{\Omega, \text{min}}}{2} s^2
	\]
	is a lower bound for all $\projstatesymb \in \Omega$.

	Moreover, for $s>1$, we consider the soft cross-entropy across $x_{\mathrm{s}}(s,\labelsymb)$,  
	
	\[
		\softlossfuncS(s,y)\coloneqq \softlossfunc(x_{\mathrm{s}}(s,\labelsymb),y)
	\]
	
	The function $\softlossfuncS$ is strictly convex since it is the soft cross-entropy $\softlossfunc$ evaluated across the line $x_{\mathrm{s}}$ along which it is strictly convex according to Lemma \ref{lemma:properties_soft_cross_entropy} and its second derivative $\frac{d^2 \softlossfuncS(s, y)}{d^2 s}>0$ strictly positive. 
	Its first derivative is zero at the minimizer, $\frac{d \softlossfuncS(s, y)}{d s}\Bigr|_{s=0}=0$, and strictly positive elsewhere since $\frac{d^2 \softlossfuncS(s, y)}{d^2 s}\Bigr|_{s\neq 0}>0$.
	Therefore, the tangent of $\softlossfuncS$  at $s=1$ $g: [0,\infty) \times \labelset \rightarrow \mathbb{R}$ is a lower bound
	\[
	\begin{aligned}
		g(s,\labelsymb) &= \softlossfuncS(1,y) + \frac{d \softlossfuncS(s, y)}{d s}\Bigr|_{s=1} (s-1) \\
		&= ms + b,
	\end{aligned}	
	\] 
	i.e., $\softlossfuncS(s,y) \geq g(s,\labelsymb)$ with strictly positive slope $m>0$. 
	
	Hence, for all $s \in \mathbb{R}^+$, the soft cross-entropy $\softlossfuncS(s,y)$ is lower bounded by $\alpha \in \mathcal{K}$ with
	\[
	\alpha(s) = 
	\begin{cases}
		\frac{\lambda_{\Omega, \text{min}}}{2} s^2 \quad & s \leq 1 \\
		m s + \lambda_{\Omega, \text{min}}/2 \quad & s > 1.
	\end{cases}
	\]	
	and $s =\lVert x_{\mathrm{s}}(s,\labelsymb) - \projstatesymb_\labelsymb^\star \rVert $.
	\PfEnd
\end{pf}

\begin{figure}
	\centering
	\setlength{\figureheight}{0.75\figurebasewidth}
	\setlength{\figurewidth}{0.85\figurebasewidth}
	\input{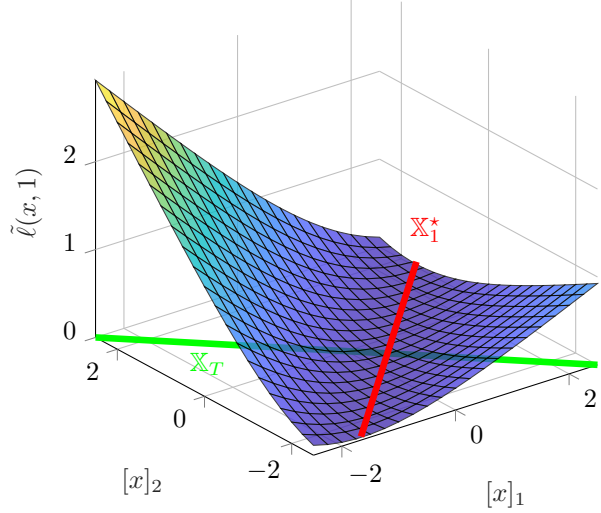}
	\caption{Illustration of the soft cross-entropy and its minimizer set for two classes with the target class $\labelsymb=1$.}
	\label{fig:soft_cross_entropy}
\end{figure}

Figure \ref{fig:soft_cross_entropy} illustrates the geometry of the soft-cross entropy  for the case of two classes.
It shows how the loss function behaves quadratic around the minimizer set before than decreasing in slope and behaving linear further away from the minimizer set.

Notice that the soft cross-entropy thus behaves similarly to the Huber loss \citep{huber1992robust} used in regression tasks, exhibiting locally quadratic and asymptotically linear behavior to reduce the influence of outliers.

\subsection{Dissipativity in ResNet Training}
The lower bound to the soft cross-entropy can now be used to formulate a strictly dissipative version of the ResNet training. 

Based on our previous analysis of the soft cross-entropy, we  now focus on the ResNet training OCP \eqref{eq:training_ocp}, when using the soft cross-entropy as a stage cost regularization
\begin{equation}
	\ell(\stateensemblesymb, \inputensemblesymb) = \offsetsoftlossfunc(\stateensemblesymb, \labelensemblesymb) + r \lVert \inputensemblesymb \rVert^2.
	\label{eq:stage_cost}
\end{equation}

\begin{prop}[Strict Dissipativity]
	\label{prop:dissipativity_soft_cross_entropy}
	The training OCP \eqref{eq:training_ocp} with the stage cost \eqref{eq:stage_cost} is strictly dissipative with respect to $\optimalsteadystateinputset=(\minimizerset_{\labelensemblesymb}, \{0\})$, with 
	\begin{equation}\label{eq:setXstar_y}
		\minimizerset_{\labelensemblesymb} = \left\{ 
		\begin{bmatrix}
			\statesymb^{1} \\
			\vdots \\
			\statesymb^{\datasetsize}
		\end{bmatrix} \in \mathbb{R}^{\datasetsize \cdot \numclassessymb} \middle| \statesymb^{d} \in \minimizerset_{\labelsymb^d}  \; \forall d = 1, \dots, \datasetsize \right\}.
	\end{equation}
	Moreover, the storage function can be chosen as $\lambda(\stateensemblesymb) = c, c \in \mathbb{R}_0^+$.
	\End
\end{prop}
	\begin{pf}
		With the ResNet dynamics \eqref{eq:resnet}, each state can be rendered a steady state for $\inputensemblesymb=0$, moreover $\inputensemblesymb=0$ is the minimizer of the input penalty $r \lVert \inputensemblesymb \rVert^2$. Therefore, the optimal steady state pairs are of the form $\left(\optimalsteadystateset, \{0\}\right)$.   
		The set of optimal steady states  $\optimalsteadystateset$ are the minimizers of the soft cross-entropy for each data sample $\left(\statesymb^d, \labelsymb^d\right)$, i.e. $ \statesymb^{d} \in \minimizerset_{\labelsymb^d}$ are the minimizers of the stage cost $\offsetsoftlossfunc(\stateensemblesymb, \labelensemblesymb)$, which correspond to $x^d$.
		The dissipation inequality with constant storage for the stacked state is the summation of the dissipation inequality for the individual data samples
		\begin{align*}
			\offsetsoftlossfunc(\stateensemblesymb, \labelensemblesymb) &= \sum_{i=1}^{\datasetsize} \offsetsoftlossfunc\left(\statesymb^i, \labelsymb^i\right)  \\ 
			&\geq \sum_{i=1}^{\datasetsize} \alpha\left(\dist \left(\statesymb^i, \minimizerset_{\labelsymb^i}\right)\right) \geq \alpha(\dist(\stateensemblesymb, \minimizerset_{\labelensemblesymb})),
			\label{eq:stage_cost_larger_distance}
		\end{align*}
		which hold according to \ref{lemma:soft_cross_entropy_distance_measure}.
		If $r > 0$, then the additional input penalty ensures strict state-input dissipativity
		$\ell(\stateensemblesymb, \inputensemblesymb) \geq \alpha(\dist((\stateensemblesymb, \inputensemblesymb), (\minimizerset_{\labelensemblesymb}, \{0\})))$. 
	\PfEnd
	\end{pf}

Table~\ref{tab:overiew_dissipativity} provides an overview of dissipativity properties for the different considered stage costs and regularizations.

\begin{table*}[h]
	\centering
	\caption{Overview of Dissipativity properties of ResNet training for different stage costs and regularizations}
	\label{tab:overiew_dissipativity}
	\begin{tabularx}{\linewidth}{p{0.2\linewidth}p{0.2\linewidth}X}
		\toprule
		Stage Cost & Strict Dissipativity & Turnpike Object \& Comments \\
		\midrule
		$\ell(\stateensemblesymb, \inputensemblesymb) =  \lVert \stateensemblesymb - \bar{\stateensemblesymb} \rVert +\lVert u \rVert^2$ & Yes & With respect to designed steady states and zero inputs, $\optimalsteadystateinputset=\left(\stateset, \left\{0\right\}\right)$  \citep{faulwasser2021turnpike}. \\
		$\ell(\stateensemblesymb, \inputensemblesymb) = \lossfunc(\stateensemblesymb, \labelensemblesymb) + \lVert u \rVert^2$ with $\lossfunc$ from \eqref{eq:cross_entropy} & Not with $\alpha_\ell \in \mathcal K_\infty$ in \eqref{eq:strict_dissipation_inequality} & Due to  turnpike set pushed to infinity.\\
		$\ell(\stateensemblesymb, \inputensemblesymb) = \softlossfunc(\stateensemblesymb, \labelensemblesymb) + \lVert u \rVert^2$ with $\softlossfunc$ from \eqref{eq:soft_cross_entropy}& Yes & With respect to the set of soft cross-entropy minimizers and zero inputs $\optimalsteadystateinputset=(\minimizerset_{\labelensemblesymb}, \{0\})$ (Proposition \ref{prop:tp_resnet}). \\
		\bottomrule
	\end{tabularx}
\end{table*}

\section{Turnpikes in ResNet Training} \label{sec:tp}

We utilize the dissipativity of the training to analyze the optimal solutions to OCP \eqref{eq:training_ocp}.

Consider the set of timesteps spent $\varepsilon$-close to the optimal state set.
\begin{equation*}
	\mathcal{Q}_\varepsilon = 
	\left\{ k \in \mathbb{N}_{[0,\networkdepthsymb-1]}  \ \middle| \ \dist\left(\stateensemblesymb_k, \optimalsteadystateset\right) \leq \varepsilon   \right\},
\end{equation*}
and its complement $\bar{\mathcal{Q}}_\varepsilon = \mathbb{N}_{[0,\networkdepthsymb-1]} \setminus \mathcal{Q}_\varepsilon$.

\begin{assum}[Exponential reachability]
	There exists constants $\rho \in [0, 1)$ and $\beta > 0$ and an infinite-horizon control input $\inputensembleinfinitesymb: \mathbb{N}_{[0, \infty)} \rightarrow \mathbb{R}^{d^2 + d}$ such that, for all initial conditions $\stateensembleinitialsymb \in \mathbf{X}_0 \subseteq \mathbb{R}^{\datasetsize \cdot \numclassessymb}$, the trajectories satisfies
	\begin{equation*}
		\dist\left(\left(\tilde{\stateensemblesymb}_k, \tilde{\inputensemblesymb}_k\right), \optimalsteadystateinputset \right) \leq \beta \rho^k,
	\end{equation*}
	\label{ass:exponential_reachability}
	where $\tilde{\stateensemblesymb}_{k+1}=\resnetensembledyn(\tilde{\stateensemblesymb}_k, \tilde{\inputensemblesymb}_k), \tilde{\stateensemblesymb}_0=\stateensembleinitialsymb$ is the corresponding state trajectory.
\End
\end{assum}
Note that while initially this assumption seems very strict, \cite{RuizBalet2021} established a similar reachability for neural ODEs using the ReLU activation function. 
Furthermore, this assumption implies a training accuracy of 100 \%, but since no assumptions are made on the test data, this could come at the price of substantial overfitting.

\begin{rem}[Exact Reachability of ResNets] \label{rem:ReachResNet}
	Consider the ResNet architecture \eqref{eq:resnet_alternative_arch} with activation functions $\sigma_1(x)=\mathrm{max}\left\{0,x\right\}$ and $\sigma_2(x)=x$.
	Let 
	\begin{equation*}
		\mathbb{D}=\left\{\left(\statesymb^1,\labelsymb^1\right), \dots, \left(\statesymb^{\datasetsize},\labelsymb^{\datasetsize}\right)\right\}
	\end{equation*} 
	be the dataset with distinct features $x^i\neq x^j, \forall i, j = 1, ..., \datasetsize$.
	Then there exists a finite horizon $\networkdepthsymb$ and parameters, $W_k$, $A_k$ and $b_k$, such that all sample from the dataset reach the respective set of soft-cross entropy minimizers, i.e. $x^i_{N} \in \optimalstateset_{\labelsymb^i} \: \forall i=1,..., \datasetsize$.
	
	This is based on the reachability result for the corresponding the neural ODE
	\[
	\dot{x}^i(t) = W(t) \sigma\left(A(t) x^i(t) + b(t)\right), ~ x^i(0) = x^i, \forall t \in [0,T],
	\]
	see Theorem 2 by  \cite{ruiz2023neural}.	
	Starting from distinct initial conditions $x^i$ and for any horizon $T$, there exists piecewise constant parameter functions $W(t), A(t), b(t)$, such that all data samples $i\in 1,..., \datasetsize$ are simultaneously controlled to their distinct targets $x^i(T) = \hat{x}^i$, with $\hat{x}^i \neq \hat{x}^j$ for $i \neq j$. 
	By choosing distinct target points $\hat{\statesymb}^i \in \optimalstateset_{\labelsymb^i}$, the neural ODE can reach the set of soft-cross entropy minimizers $x^i(T) \in \optimalstateset_{\labelsymb^i}$ with piecewise constant parameter functions $W(t), A(t), b(t)$ and at most $6\cdot \datasetsize$ switches.
	By choosing a sufficiently small discretization time $h> 0$, the results transfer to the ResNet on the horizon $N = \left\lceil \frac{T}{h}\right\rceil$, for details we refer to \cite[Remark 3.2]{ruiz2023neural}.
	\End
\end{rem}

\begin{prop}[Turnpikes in ResNet Training]
	\label{prop:tp_resnet}
	Consider the training OCP \eqref{eq:training_ocp} with stage cost \eqref{eq:stage_cost}.
	Suppose that Assumption \eqref{ass:exponential_reachability} holds.
	Then, there exists a constant $\hat{V}$ such that, for any $\gamma\in \mathbb{R}$ in OCP \eqref{eq:training_ocp}, the optimal solutions satisfy
	\begin{equation}
		\# Q_{\varepsilon} \geq \networkdepthsymb-\frac{\hat{V}}{\alpha(\varepsilon)} \qquad\qquad  \# \hat{Q}_{\varepsilon} \leq \frac{\hat{V}}{\alpha(\varepsilon)},
		\label{eq:cardinality_turnpike}
	\end{equation}
	where $\#\mathcal{Q}_\varepsilon$ is the cardinality of the set $\mathcal{Q}_\varepsilon$.
	\End
	\end{prop}
	\begin{pf}
		The soft cross-entropy loss function is globally Lipschitz with constant $L$, i.e. for all $\stateensemblesymb_1, \stateensemblesymb_2 \in  \mathbb{R}^C$ it holds that
		\[
		\begin{split}
			\left| \offsetsoftlossfunc(\stateensemblesymb_1, \labelensemblesymb)-  \offsetsoftlossfunc(\stateensemblesymb_2, \labelensemblesymb) \right| \leq L \lVert \stateensemblesymb_1 - \stateensemblesymb_2 rVert| \\
			\leq \max_{\tilde{\mathbf{x}} \in \mathbb{R}^{\datasetsize \cdot \numclassessymb}} \lVert \nabla_{\tilde{\mathbf{x}}}  \offsetsoftlossfunc(\tilde{\mathbf{x}}, \labelensemblesymb) \rVert \lVert \stateensemblesymb_1 - \stateensemblesymb_2 \rVert.
		\end{split}
		\] 
		The gradient is bounded because of the softmax $p(x) \in [0,1]^\numclassessymb$ and target probabilities  $\tilde{q}(y)  \in [0,1]^\numclassessymb$
		\[
		\begin{split}
			\max_{\tilde{\stateensemblesymb} \in \mathbb{R}^{\datasetsize \cdot \numclassessymb}} \lVert \nabla_{\tilde{\mathbf{x}}}  \offsetsoftlossfunc(\tilde{\stateensemblesymb}, \labelensemblesymb) \rVert 
			\leq \max_{\tilde{\statesymb} \in \mathbb{R}^{\numclassessymb}} \lVert| \nabla_{\tilde{\statesymb}}  \offsetsoftlossfunc(\tilde{\statesymb}, \labelsymb) \rVert \\
			= \max_{\tilde{\statesymb} \in \mathbb{R}^{ \numclassessymb}} \lVert p(\tilde{\statesymb}) - \tilde{q}(x) \rVert \leq  \max_{\tilde{\statesymb} \in \mathbb{R}^{\numclassessymb}} \lVert p(\tilde{\statesymb}) \rVert  + \lVert \tilde{q} (y) \rVert \\
			\leq \numclassessymb + \numclassessymb = 2 \numclassessymb.
		\end{split}
		\]
		
		Using the Lipschitz constant, Assumption \ref{ass:exponential_reachability} implies that there exists $\hat{V}$ such that $V^\gamma_\networkdepthsymb \left(\stateensembleinitialsymb \right) \leq \hat{V}$
		\[
		\begin{aligned}
						V^\gamma_\networkdepthsymb \left(\stateensembleinitialsymb \right) &=
			\sum_{k = 0}^{\networkdepthsymb-1}
			\left[\offsetsoftlossfunc(\stateensemblesymb_k, \labelensemblesymb) + r \lVert \inputensemblesymb \rVert^2 \right] + \gamma \offsetsoftlossfunc(\stateensemblesymb_\networkdepthsymb, \labelensemblesymb) \\
			&\leq  
			\sum_{k = 0}^{\networkdepthsymb-1}
			L \dist\left(\left(\tilde{\stateensemblesymb}_k, \tilde{\inputensemblesymb}_k\right), \optimalsteadystateinputset \right) +  L \dist\left(\left(\tilde{\stateensemblesymb}_\networkdepthsymb, \tilde{\inputensemblesymb}_\networkdepthsymb\right), \optimalsteadystateinputset \right) \\
			& \leq 
			L \left[
						\sum_{k = 0}^{\networkdepthsymb-1}
			\beta \rho^k + \gamma \beta \rho^N
			\right] \leq 
			\left[ L \beta \frac{1}{1-\rho} + \beta \right] = \hat{V}.
		\end{aligned}
		\]

		Then due to Proposition \ref{prop:dissipativity_soft_cross_entropy}
		\begin{equation*}
			\begin{aligned}
				\storagefuncresnet(\stateensemblesymb^\star_\networkdepthsymb) - \storagefuncresnet(\stateensemblesymb_0)& \leq \sum_{k=0}^{\networkdepthsymb-1} \ell(\stateensemblesymb^\star_k, \inputensemblesymb^\star_k) - \alpha(\dist(\stateinputensemblesymb^\star_k, \optimalsteadystateinputset))    \\
				\Leftrightarrow  
				\storagefuncresnet(\stateensemblesymb^\star_\networkdepthsymb) - \storagefuncresnet(\stateensemblesymb_0)& +
				\sum_{k=0}^{\networkdepthsymb-1} \alpha(\dist(\stateinputensemblesymb^\star_k, \optimalsteadystateinputset))\\ &\leq 
				V_\networkdepthsymb^{\gamma} \left(\stateensembleinitialsymb\right) -\gamma \lossfunc(\stateinputensemblesymb^\star_\networkdepthsymb, \labelensemblesymb).
			\end{aligned}
		\end{equation*}
		Due to the constant storage dissipativity $\lambda(\stateensemblesymb)=c, c\in\mathbb{R}^+_0$, $\storagefuncresnet(\stateensemblesymb^\star_N)- \storagefuncresnet(\stateensemblesymb_0)=0$.
		Therefore
		\begin{equation*}
			(N-\#\mathcal{Q}_\varepsilon) \alpha_\ell(\varepsilon) \leq \sum_{k=0}^{\networkdepthsymb-1} \alpha_\ell(\dist(\stateinputensemblesymb^\star_k, \optimalsteadystateinputset)) \leq \hat{V}.
		\end{equation*}
		Rearranging gives \eqref{eq:cardinality_turnpike}.
	\PfEnd
	\end{pf}

For a singleton optimal steady state-input set $\optimalsteadystateinputset = \left\{ \optimalsteadystate; \optimalsteadyinput \right\}$, this corresponds to Proposition 2 from \cite{faulwasser2021turnpike} or to similar results by \cite{Gruene13a} for generic discrete-time optimal control problems.

Here, however, the turnpike $\optimalsteadystateinputset = (\minimizerset_\labelensemblesymb, \left\{0\right\})$ for the stacked data samples consists of individual subspace turnpikes for each class $\minimizerset_1,\dots,\minimizerset_\numclassessymb$ to which the trajectories of samples belonging to the respective class converge.

\section{Neural ODEs and Other NN Architectures}
\label{sec:neuralODEs}
Our preceding analysis has focused on classic ResNets which we conceptualize a discrete-time systems. However, in view of the link between neural ODEs \eqref{eq:NODE} and ResNets \eqref{eq:resnet}, we first turn towards the continuous-time extension and discuss other architectures. 

\subsection{Continuous-Time Training Formulation}
The neural ODE counterpart to \eqref{eq:training_ocp} reads
\begin{subequations}\label{eq:ct_training_ocp}
	\begin{align}
		\min_{\inputensemblesymb(\cdot) \in \mathcal{L}^\infty}& \int_{0}^{T} \ell(\stateensemblesymb(t), \inputensemblesymb(t)) \mathrm{d}t+ \gamma\lossfunc(\stateensemblesymb(T), \labelensemblesymb)\\
		\text{s.t.}&\notag \\
		 \quad \quad& \dot{\stateensemblesymb}(t) =
\mathrm{f}_\mathrm{c}(\stateensemblesymb(t), \inputensemblesymb(t)), \\ \quad \quad&\stateensemblesymb(0) =  \stateensembleinitialsymb \in \mathbb{R}^{\datapointdimsymb \cdot \datasetsize},
	\end{align}
\end{subequations}
where the input signal stacks the  bias and weight functions $A(t), b(t)$ and is considered to be in $\mathcal{L}^\infty([0,T], \mathbb R^{n_u})$ and the same data-stacking procedure as in Section~\ref{sec:background} is applied.  The continuous time variable $t$ models the network depth and the stacked dynamics are
\begin{equation} \label{eq:NODEstacked}
\begin{aligned}
\dot{\stateensemblesymb}(t) =& 
\mathrm{f}_\mathrm{c}(\stateensemblesymb(t), \inputensemblesymb(t)) \\
	\doteq&	
			 \sigma \left((\identitymat{\datasetsize} \otimes A(t))\stateensemblesymb(t)+(\vectone{\datasetsize} \otimes b(t))\right).
\end{aligned}
\end{equation}

The counterpart to the set $\bar{\stateinputset}$ from \eqref{eq:Xs}
reads
\begin{equation}
 \bar{\stateinputset}_\mathrm{c} = \left\{(\bar{\stateensemblesymb}, \bar{\inputensemblesymb}) \in \mathbb{R}^{\datasetsize\cdot \numclassessymb}\times \mathbb{R}^{(C^2+C)N} \,|\,0=\mathrm{f}_\mathrm{c}(\bar{\stateensemblesymb}, \bar{\inputensemblesymb}) )
\right\}
\end{equation}
Similar to before, optimal steady state pairs are computed via
	\begin{equation}
		\begin{aligned}
			\bar{\stateinputensemblesymb}^\star \in
			\argmin_{\bar{\stateinputensemblesymb} } 
			&\; \ell(\bar{\stateinputensemblesymb}) \quad 
			 \text{s.t.} \quad \bar{\stateinputensemblesymb} \in \steadystateinputset_\mathrm{c} 
		\end{aligned}
	\end{equation}
	and the set of all optimal steady states is written as $\optimalsteadystateinputset_\mathrm{c} \subseteq \bar{\stateinputset}_\mathrm{c}$.

In view of Definition~\ref{def:DI} continuous-time OCPs are called strictly dissipative with respect to $\optimalsteadystateinputset_\mathrm{c}$.
\begin{defn}[Strict dissipativity  in cont. time] \label{def:DI_ct}
	\begin{subequations} \label{eq:ct_dissipation_inequality}
		The dynamical system \eqref{eq:NODEstacked} is said to be dissipative with respect to a set of steady-state pairs $\optimalsteadystateinputset_\mathrm{c}$ if there exists a non-negative storage function $\lambda: \stateset \rightarrow \mathbb{R}^+_0$ such that for all pairs $\mathbf{z}(t)=(\stateensemblesymb(t),\inputensemblesymb(t))$ defined over some interval $[0,T]$ and all $\bar{\stateinputensemblesymb}^\star \in \optimalsteadystateinputset_\mathrm{c}$ 
		\begin{equation}
			\lambda(\stateensemblesymb(T)) - \lambda(\stateensemblesymb(0)) \leq \int_0^T \ell(\stateinputensemblesymb(t)) - \ell(\bar{\stateinputensemblesymb}^\star)\mathrm d t.	
		\end{equation}
				If additionally, there exists $\alpha_{\ell}  \in \mathcal{K}$ such that 
		\begin{multline}
				\lambda(\stateensemblesymb(T)) - \lambda(\stateensemblesymb(0)) \\ \leq \int_0^T \ell(\stateinputensemblesymb(t)) - \ell(\bar{\stateinputensemblesymb}^\star)		- \alpha_{\ell}\left(
			\mathrm{dist}\left(\stateinputensemblesymb(t), \optimalsteadystateinputset\right)\right)\mathrm d t,
			\label{eq:_ctstrict_dissipation_inequality_c}
		\end{multline}
					\end{subequations}
		then the system \eqref{eq:NODEstacked} is said to be strictly $\stateensemblesymb-\inputensemblesymb$ dissipative with respect to $\optimalsteadystateinputset_\mathrm{c}$ and for $\stateinputensemblesymb(t)$ replaced by $\stateensemblesymb(t)$ in the class $\mathcal K$ function the system is said to be strictly $\stateensemblesymb$ dissipative. 
		
			The OCP \eqref{eq:ct_training_ocp} is said to be (strictly) $\stateensemblesymb-\inputensemblesymb$ dissipative with respect to $\optimalsteadystateinputset_\mathrm{c}$ if for all $T\in\mathbb{R}_0^+$ and all $\stateensemblesymb_0 \in \mathbf{X}_0$, the dissipation inequalities \eqref{eq:ct_dissipation_inequality} hold along any optimal trajectory of \eqref{eq:ct_training_ocp}.
\End
\end{defn}

\subsection{Continuous-Time Results}

The following corollary to Proposition \ref{prop:dissipativity_soft_cross_entropy} holds in the continuous-time setting. It follows from the observations that the geometry of the soft entropy loss function is not altered in the continuous-time setting and that again the entire state space of \eqref{eq:NODEstacked} is covered by steady states with $\bar u =0$.

\begin{cor}	\label{cor:dissipativity_soft_cross_entropy}
	Consider the training OCP \eqref{eq:ct_training_ocp} with the stage cost \eqref{eq:stage_cost}. 
	Then the OCP is strictly dissipative with constant storage $\lambda(\stateensemblesymb) = c, c\in \mathbb R^+_0$ and with respect to $\optimalsteadystateinputset_\mathrm{c}=(\minimizerset_{\labelensemblesymb}, \{0\})$ with $\minimizerset_{\labelensemblesymb}$ from \eqref{eq:setXstar_y}.
	\End
\end{cor}

The set 
\begin{equation*}
\Theta_{\varepsilon, T}(x_0) \doteq \left\{t \in [0,T]\,|\, \dist\left(\left({\stateensemblesymb}^\star(t), \ {\inputensemblesymb}^\star(t)\right), \optimalsteadystateinputset_\mathrm{c} \right)> \varepsilon \right\}
\end{equation*}
collects all time points for which the optimal pairs stay outside of an $\varepsilon$-neighbourhood of $\optimalsteadystateinputset_\mathrm{c}$.
The reachability property used for the corresponding turnpike results is stated in the next assumption. 
\begin{assum}[Exp. reachability in cont. time]	\label{ass:ct_exponential_reachability}
	There exists constants $\rho >0 $ and $\beta > 0$ and an infinite-horizon control input $\tilde{\inputensemblesymb}(\cdot) \in \mathcal{L}^\infty([0,T], \mathbb R^{n_u})$ such that, for all initial conditions $\stateensembleinitialsymb \in \mathbf{X}_0 \subseteq \mathbb{X}$, the trajectories of \eqref{eq:NODEstacked} satisfy
	\begin{equation*}
		\dist\left(\left(\tilde{\stateensemblesymb}(t), \tilde{\inputensemblesymb}(t)\right), \optimalsteadystateinputset_\mathrm{c} \right) \leq \beta \exp(-\rho t),
	\end{equation*}
	where $\tilde{\stateensemblesymb}(t)$ is the corresponding state trajectory driven by $\inputensembleinfinitesymb(\cdot)$.
\End
\end{assum}
The next result translates Proposition \ref{prop:tp_resnet} to the continuous-time neural ODE setting. 
\begin{prop}[Measure Turnpikes]
	Consider the training OCP \eqref{eq:ct_training_ocp} with stage cost \eqref{eq:stage_cost}.
	Suppose that \eqref{ass:ct_exponential_reachability} holds.
	Then, there exists a continuous function $\nu:(0,\infty] \to \mathbb R^+_0$ independent of $T$  such that, for any $\gamma\in \mathbb{R}$ in  OCP \eqref{eq:training_ocp}, the optimal solutions satisfy
	\begin{equation}
		\mu[\Theta_{\varepsilon, T}(x_0)] \leq \nu(\varepsilon)< \infty,
		\label{eq:measure_turnpike}
	\end{equation}
	where $\mu$ is the Lebesgue measure on the real line.
	\End
	\end{prop}
	The proof follows the usual structure of dissipativity-based proofs of measure turnpikes~\citep{carlson91nfinite,epfl:faulwasser15h}. It is thus omitted. 
	
	\subsection{Extension to Other NN Architectures}
	
Our results so far raise the question of whether or not one could consider other architectures than ResNets?
To this end, we first formalize the relation between the neural ODEs and the considered NN architecture. 
\begin{defn}[Equilib. consistent discretization]
Consider a continuous-time system of the form 
\[
0=  F_\mathrm{c}(\dot x,x,u), \quad x(0) = x^0 \in\mathbb R^C.
\]
Its discretization
\[
0=  F_\mathrm{d}(x_{k+1},x_k,u_k), \quad x_0 = x^0 \in\mathbb R^C
\]
is said to preserve the equilibria if any $(\bar x, \bar u)$ which solves $0=  F_\mathrm{c}(0,\bar x,\bar u)$ also solves
$0=  F_\mathrm{d}(\bar x,\bar x,\bar u)$ and vice-versa.\End
\end{defn}

\begin{prop}[Dissipativity \& architectures] \label{prop:DI_Architecture}
Consider the discrete-time training OCP \eqref{eq:training_ocp} with the soft-entropy stage cost \eqref{eq:stage_cost} whereby the considered NN architecture captured in \eqref{eq:training_ocp_dyn} is an equilibria preserving discretization of \eqref{eq:NODEstacked}. 
	Then OCP \eqref{eq:training_ocp} is strictly dissipative with respect to $\optimalsteadystateinputset=(\minimizerset_{\labelensemblesymb}, \{0\})$ from \eqref{eq:setXstar_y}
			and the storage function can be chosen as $\lambda(\stateensemblesymb) = c, c \in \mathbb{R}_0^+$.
\end{prop}
\begin{pf}
The proof follows from the observation that the structure of the neural ODE \eqref{eq:NODEstacked} combined with $\sigma(0) = 0$ implies that for $u=0$ the entire state space of \eqref{eq:NODEstacked} is covered by equilibria. Then applying any equilibria preserving discretization (which may be implicit or explicit, fixed step size of variable step size) means that also in  OCP \eqref{eq:training_ocp} the dynamics \eqref{eq:training_ocp_dyn} are such that the entire state space consist of equilibria corresponding to $u =0$. Hence the analysis of the geometry of the soft cross-entropy stage cost \eqref{eq:stage_cost} can be conducted as in Lemma~\ref{lemma:properties_soft_cross_entropy}. \PfEnd
\end{pf}

The previous result shows that the crucial requirement on the network architecture is that the entire state space is covered by equilibrium points for zero inputs (which depends on the architecture and the activation function). Thus, our dissipativity results readily transfer to other ResNet architectures consisting of multiple layers with one skip connection \citep{he2016deep,ESTEVEYAGUE2023105452}, e.g., the architecture
\begin{equation}
	\begin{aligned}
		\statesymb^i_{k+1} &= \statesymb^i_k + \sigma_2\big(A_{k, 2}\sigma_1\left(A_{k, 1} \statesymb^i_k+b_{k, 1}\right)+b_{k, 2}\big) \\
		\statesymb^i_0&=\statesymb^i \in \mathbb{R}^{\datapointdimsymb},
	\end{aligned}
	\label{eq:resnet_alternative_arch}
\end{equation}
which first decreases the state dimension to a hidden dimension $h$ by $A_{k,1} \in \mathbb{R}^{h \times C}, b_{k,1} \in \mathbb{R}^{h}$ and then increases it again by $A_{k,2} \in \mathbb{R}^{C \times h}, b_{k,2} \in \mathbb{R}^{C}$. It thus offers fewer trainable parameters per layer and non-linearity \citep{esteve2021largetime}. Turnpike results similar to Proposition \ref{prop:tp_resnet} are readily inferred if the setting of  Proposition~\ref{prop:DI_Architecture} is combined with suitable reachability properties. In view of Remark \ref{rem:ReachResNet} notice that sufficiently accurate equilibria preserving discretizations also preserve reachability, cf. the results of \cite{ruiz2023neural}.

\begin{rem}[Link to deep equilibrium networks]
NN architectures can be obtained from the discretization of neural ODEs or from other considerations. 
Deep equilibrium networks, e.g., are a 
recently proposed approach built around the observation that in many cases the hidden layers of ANNs approach a steady state before this steady state is propagated through an output layer \citep{bai2019deep,ling2024deep}. 
The core idea of equilibrium networks is to consider the implicit steady equation as the model of data propagation. That is, these networks directly solve for the unknown network equilibrium, e.g., via tailored variants of Newton's method. 

In contrast, in the present paper, and in different  fashion also in \citep{faulwasser2021turnpike},  we shift finding the equilibrium to the training, while the optimal equilibrium subspace (the present paper) or a chosen pre-computed equilibrium \citep{faulwasser2021turnpike} for the loss function  is encoded in the 
regularization stage cost. In depth exploration of the links between deep equilibrium networks and our approach is subject to future work.
\end{rem}

	\section{Numerical Experiments}
\label{sec:experiments}

To validate and illustrate the dissipative formulation of the ResNet training with soft-cross entropy we train networks on the two-spirals task and on the MNIST dataset \citep{deng2012mnist}. 
The training is implemented in Python using the PyTorch framework for NNs \citep{pytorch2019}. 

\begin{figure}
	\centering
	\includegraphics[width=\figurebasewidth]{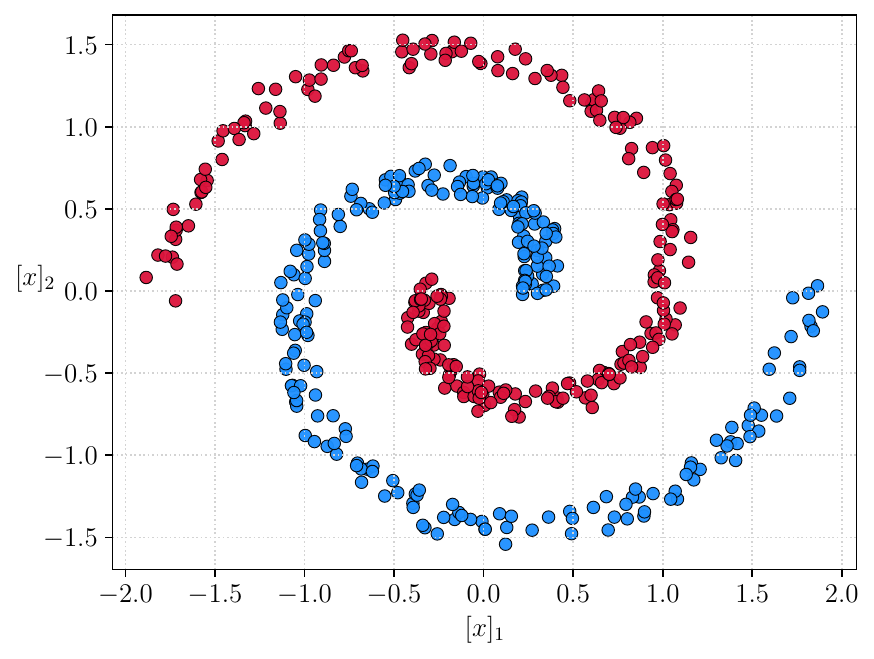}
	\caption{Two Spirals dataset.}
	\label{fig:two_spirals}
\end{figure}

As a first experiment, we consider the two-spirals task classification problem  of separating two intertwined spirals~\citep{twospirals}. 
The dataset comprising 480 training samples is visualized in Figure \ref{fig:two_spirals}.
On the dataset, we train a 30-layer ResNet with the architecture \eqref{eq:resnet_alternative_arch} with a hidden dimension of $h=8$, a hyperbolic activation function $\sigma_1(x) =\tanh(x)$ and one identity activation function $\sigma_2(x)=x$. 
For the label smoothing we use $p_d=0.95$ as the probability for the correct class.
The network is trained using the Adam optimizer \citep{adam} with a learning rate $\alpha=0.1$ and weight decay $r=0.005$ and a terminal penalty $\gamma =3$. 

\begin{figure}
	\centering
	\includegraphics[width=\figurebasewidth]{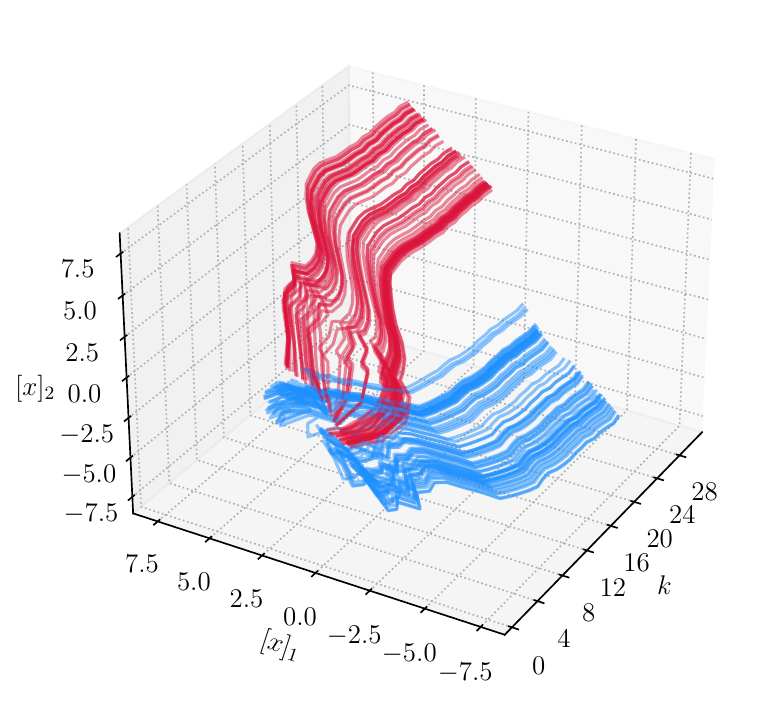}
	\caption{Evolution of the state trajectories for the two classes of the two spirals dataset.}
	\label{fig:two_spirals_evolution}
\end{figure}

\begin{figure}
	\centering
	\includegraphics[width=\figurebasewidth]{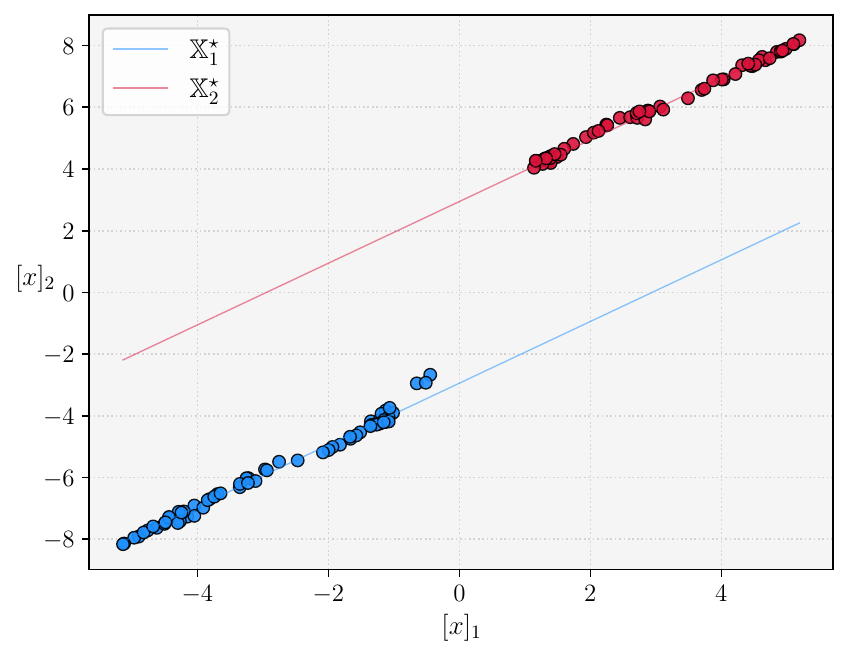}
	\caption{State of the data trajectories int he last layer and the sets of soft-cross entropy minimizers for the two classes, $\minimizerset_{1}$ and $\minimizerset_{2}$.}
	\label{fig:two_spirals_last_layer}
\end{figure}

The evolution of data trajectories in Figure \ref{fig:two_spirals_evolution} shows how the two classes are separated in the first ten layers, after which the optimal steady state is reached and the data exhibits the turnpike phenomenon.
In addition, the final layer's data trajectories, as shown in Figure \ref{fig:two_spirals_last_layer}, lie closely scattered around the minimizer sets for both classes.

To analyze the turnpikes for real-world datasets we consider the MNIST dataset for classifying handwritten digits from $0$ to $9$ as a second experiment. 
Each $28\times 28$ pixel image and is flattened to create the corresponding feature $x^i\in \mathbb{R}^{784}$. 
We use the ResNet architecture \eqref{eq:resnet_alternative_arch} with a hidden dimension of $h=128$ and one ReLu $\sigma_1(x) = \mathrm{max}\left\{0,x\right\}$ and one identity activation function $\sigma_2(x)=x$ and train it according to the OCP \eqref{eq:training_ocp} with $\gamma=1$ and weight decay $r=10^{-5}$. 
For the label smoothing we use $p_d=0.91$ as the probability for the correct class.
The PyTorch optimization hyperparameters are tuned to minimize the training loss, i.e., to foster the visibility of the turnpike phenomenon.
The network depth is choosen large to obtain turnpikes.

The softmax probabilties and therefore the loss function are calculated from the first 10 components of each data sample, see Remark \ref{remark:n_new_c}.
Although the remaining states are not considered in the loss calculation, the weight matrices of the ResNet can still utilize the information contained in these states to aid the classification in the first ten states.

\begin{figure}
	\centering
	\includegraphics[width=\figurebasewidth]{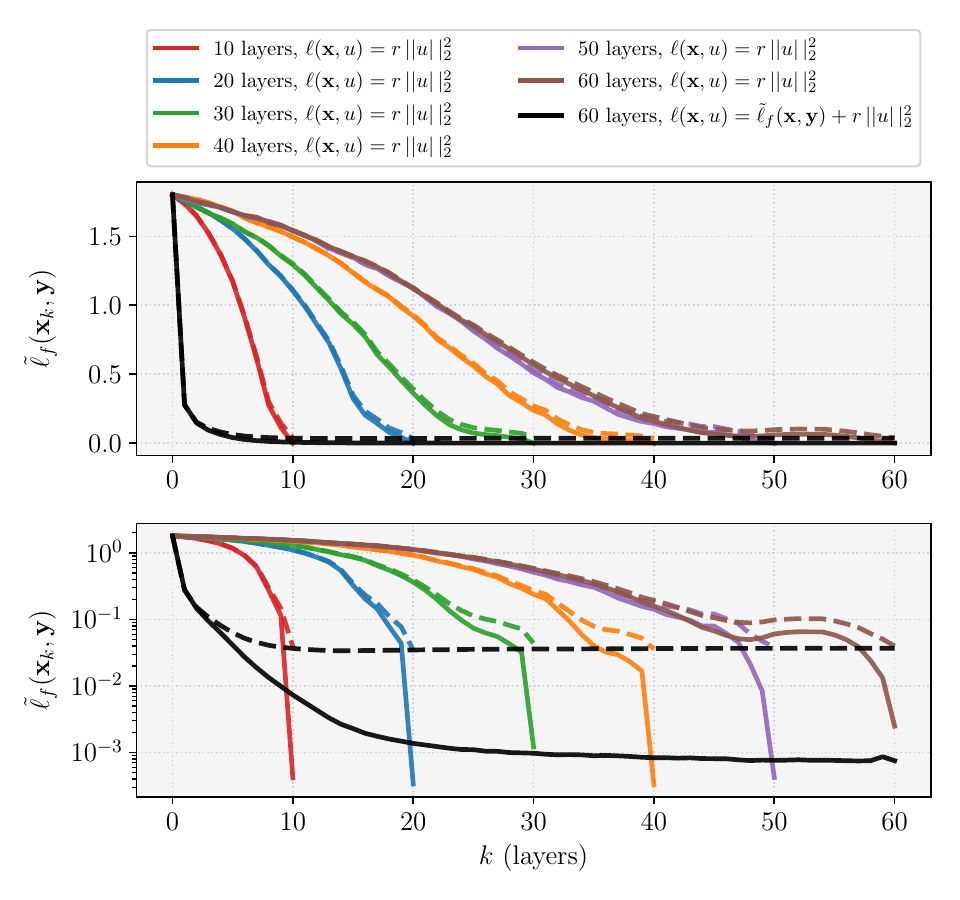}
	\caption{The loss over the layers of the ResNet for the MNIST dataset in linear and logarithmic scale. The straight line represents the training loss and the dashed line represents the test loss.}
	\label{fig:mnist}
\end{figure}

Figure \ref{fig:mnist} shows a comparison of training with and without stage cost.
It shows the evolution of the training loss over the layers of the ResNet for a 60-layer network trained with the soft-cross entropy in stage cost in comparison to networks trained without the soft cross-entropy stage cost with depths from 10 to 60 layers.
The networked trained with the stage cost takes around 10 layers to reduce the stage cost and to achieve a good classification result. After that the stage cost remains at the low level indicating the turnpike phenomenon.
Meanwhile, when training with input penalization only, the network always needs its full depth to achieve the classification tasks.
The turnpike phenomenon thus allows to crop the turnpike layers which thus do not contribute to the transformation learned.
The dissipative formulation thus allows determining the depth required for a classification task without additional  hyperparameter tuning for the depth of the network. Moreover, the proposed training formulation allows to start with conservative guesses for the required network depth. If the guess turns out to be too conservative, the trained networked can be cropped and used, alternatively it has to be retrained with more neurons.

	\section{Conclusions}
\label{sec:conclusion}

This work has formulated a dissipative version of ResNet training for classification wherein a regularization based on a variant of the cross-entropy in the stage cost is used. 
Assuming asymptotic reachability of the set of minimizers, we prove the existence of turnpikes in the training formulated as an optimal control problem. This phenomenon allows  to simplify the tuning of the network depth, since the last layers of the network do not contribute to the learned transformation and can be removed without changing the performance of the network. We have also discussed the extension to neural ODEs and to other NN architectures with skip connections.
Experiments on the simple two-spirals testcase and on the MNIST dataset validate the turnpike results and show that they also extend to the test data.

Future work will consider the extension to more general neural network architectures and the analysis of generalization properties of the trained neural network. Moreover, a suboptimality analysis, i.e. an analysis which quantifies the required degree of optimality to observe the turnpike phenomenon, should be conducted.

	\begin{acknowledgements}
		This work has been partly funded by the Federal Ministry of Education and Research (BMBF) via the project 6GEM under funding reference 16KISK038.
	\end{acknowledgements}
	
	\printbibliography

\end{document}